\documentclass[preprint,12pt]{elsarticle}
\usepackage{amsmath,amsfonts}
\usepackage{algorithmic}
\usepackage{algorithm}
\usepackage{array}
\usepackage[caption=false,font=normalsize,labelfont=sf,textfont=sf]{subfig}
\usepackage{textcomp}
\usepackage{stfloats}
\usepackage{url}
\usepackage{verbatim}
\usepackage{graphicx}
\usepackage{mathtools}
\usepackage{multirow}
\usepackage{parskip}
\usepackage{times}
\usepackage[caption=false]{subfig}
\usepackage{dirtytalk}

\usepackage{xcolor}

\usepackage{amssymb}

\journal{ArXiv}

\begin{document}

\begin{frontmatter}




\title{Fusion Flow-enhanced Graph Pooling Residual Networks for Unmanned Aerial Vehicles Surveillance in Day and Night Dual Visions}
\author{Alam Noor \fnref{myfootnote1}}
\author{Kai Li \fnref{myfootnote1}}
\ead{kaili@ieee.org}
\cortext[mycorrespondingauthor]{Corresponding author}
\author{Eduardo Tovar \fnref{myfootnote1}}
\author{Pei Zhang \fnref{myfootnote2}}
\author{Bo Wei \fnref{myfootnote3}}

\address{1. CISTER Research Center, Porto, Portugal}
\address{2. University of Michigan, Ann Arbor, Michigan, USA.}
\address{3. Newcastle University, Newcastle, UK.}
\fntext[myfootnote1]{CISTER Research Center, Porto, Portugal}
\fntext[myfootnote2]{University of Michigan, Ann Arbor, Michigan, USA.}
\fntext[myfootnote3]{Newcastle University, Newcastle, UK.}
\begin{abstract}
Recognizing unauthorized Unmanned Aerial Vehicles (UAVs) within designated no-fly zones throughout the day and night is of paramount importance, where the unauthorized UAVs pose a substantial threat to both civil and military aviation safety. However, recognizing UAVs day and night with dual-vision cameras is nontrivial, since red-green-blue (RGB) images suffer from a low detection rate under an insufficient light condition, such as on cloudy or stormy days, while black-and-white infrared (IR) images struggle to capture UAVs that overlap with the background at night. In this paper, we propose a new optical flow-assisted graph-pooling residual network (OF-GPRN), which significantly enhances the UAV detection rate in day and night dual visions. The proposed OF-GPRN develops a new optical fusion to remove superfluous backgrounds, which improves RGB/IR imaging clarity. Furthermore, OF-GPRN extends optical fusion by incorporating a graph residual split attention network and a feature pyramid, which refines the perception of UAVs, leading to a higher success rate in UAV detection. 
A comprehensive performance evaluation is conducted using a benchmark UAV catch dataset. The results indicate that the proposed OF-GPRN elevates the UAV mean average precision (mAP) detection rate to 87.8\%, marking a 17.9\% advancement compared to the residual graph neural network (ResGCN)-based approach.
\end{abstract}





\end{frontmatter}


\section{Introduction}
\begin{figure}[htbp!]
    \begin{center}
    \hspace*{0em}
        \includegraphics[scale=0.091]{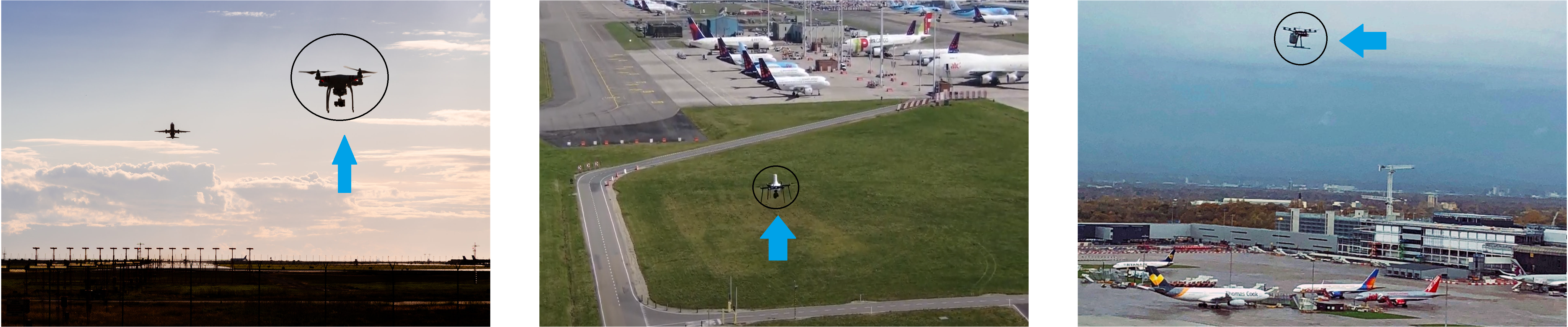}
        \caption{Unauthorized UAVs patrol in the no-fly zone, which disturbs the takeoff and landing of the flight or spies on confidential military operations.}
        \label{UAVsUses}
    \end{center}
\end{figure}

Detecting illegal unmanned aerial vehicles (UAVs) within designated no-fly zones is of paramount importance, where the illegal UAVs pose a substantial threat to civil and military aviation safety due to their potential to interfere with flight paths, causing severe accidents \cite{8447902}. The illegal UAVs also present a risk to sensitive infrastructure, such as power plants and communication networks \cite{8601408, 10246260}, where an accidental or intentional collision could result in widespread service disruptions or catastrophic failures \cite{8960477}. As shown in Fig.~\ref{UAVsUses}, an unauthorized UAV outfitted with cameras or other surveillance devices in a no-fly zone can infringe on privacy rights and present substantial security threats to both civil and military operations. Such UAVs have the potential to obtain unauthorized imagery or data, thereby providing malicious entities with invaluable information \cite{9778246}. 

Identifying UAVs through camera imagery presents a considerable challenge. This arises from UAVs' propensity to integrate inconspicuously with environmental elements, such as structures and foliage, particularly during nocturnal hours when their hues can closely resemble the backdrop. Moreover, during daylight, the variability in illumination conditions further complicates the detection process.

Deep learning models, referenced in \cite{9519550, https://arxiv.org/abs/2209.03625, 9578486, uavbenchmark}, have been utilized for UAV detection based on their color congruence with the background. Notably, these models often demonstrate proficiency in detecting UAVs in daytime color images or in infrared (IR) images captured at night. {However, its efficacy tends to diminish when faced with homogenous backgrounds at nighttime or fluctuating illumination during daylight hours.

In our antecedent research \cite{9750351}, we explored a deep learning model, leveraging transformations and cosine annealing strategies to reduce classification and regression discrepancies for UAV detection utilizing both RGB and IR imagery. However, the detection efficacy using RGB images is compromised under less-than-ideal lighting conditions, such as during overcast or tempestuous days. On the other hand, while using IR (monochromatic) imagery, discerning UAVs that merge with backgrounds of a similar hue presents its own set of challenges.

In this paper, we propose optical flow-assisted graph pooling residual networks (OF-GPRN) designed for intricate UAV detection using combined RGB and IR images. While the IR image remains unaffected by light conditions, the RGB image retains vital color data. The proposed OF-GPRN takes advantage of integrating RGB and IR images to optimize contrast, edge definition, color, and texture in each frame. This combined image also mitigates distortions arising from lighting variances, color, and background interference. Relying on this image integration, the OF-GPRN system produces a comprehensive composite frame enriched with features such as fine-texture, broad-texture, and contrast, which facilitates the extraction of UAV movement patterns. 

To isolate the UAV from its background, our proposed OF-GPRN system harnesses the fusion of RGB and IR images, subsequently processed through optical flow \cite{9495134}, facilitating the segregation of the UAV from the amalgamated image. This system innovatively augments graph neural networks (GCN) by integrating graph residual split-attention networks (GRSaN) \cite{9408381,9008535,9793853}, aiming to optimize the mAP for UAV detection \cite{zhang2020resnest}. Given the diminutive representation of the extracted UAV within the image, it poses challenges in distinguishing it from other entities, such as avian creatures or aircraft. Specifically, the OF-GPRN model refines the extracted object's contours and ascertains pixel correlations across the pre-processed RGB and IR imagery. This aids in UAV identification and augments predictive accuracy by capitalizing on the expansive feature-learning prowess offered by the feature pyramid during model calibration. 

The main contributions of this paper are listed as follows:
\begin{itemize}
    \item The OF-GPRN system is proposed to enable precise UAV detection in day and night dual visions that experience time-varying lighting conditions and high background similarities. The OF-GPRN system develops a fusion of RGB and IR frames, which enhances the quality of output frames by reducing noise and adjusting illumination and color. The OF-GPRN system also extends a new optical flow model to eliminate background and foreground similarity while extracting the UAV’s mobility.
    \item The GRSaN is extended in the OF-GPRN system to stabilize the learning capability, enhance feature learning during training, and reshape the UAV. The OF-GPRN system also uses a Quickshift-based algorithm to represent the adjacency matrix from pixels to nodes. This makes for an accurate graph with clear frame-pixel information.
    \item We conducted experiments to assess the performance of the proposed OF-GPRN system in UAV detection during both daytime and nighttime conditions. In comparison to preceding residual GCN (ResGCN) object detectors, our enhanced model demonstrates superior performance, achieving a commendable mAP of 87.8\% on the stringent RGB-IR combined benchmark UAV catch dataset. This marks a significant improvement over the ResGCN, which obtains an mAP of 69.9\%. 
\end{itemize}

This paper is organized as follows:
Section \ref{Related work} presents the literature overview on deep-learning-based UAV detection. The proposed OF-GPRN system is presented in Section \ref{Methodology}. In Section \ref{Experiments}, we study the system implementation, experimental setup, as well as the performance evaluation. Section \ref{Conclusion} concludes the paper. The symbols used in the paper have been listed below in the Table. \ref{Notations}.
\begin{table}[htbp!]
\caption{Basic Notations}
    \label{Notations}
    \hspace*{3em}
\begin{tabular}{ |p{4cm}||p{6cm}|}
\hline
\textbf{Symbol} & \textbf{Definition} \\
\hline
\hline
$\mathcal{L}_{(x,y)}$ & Modified Laplacian operator.\\
\hline
${L\;}_{(x,y)}$ & Modified Laplacian. \\
\hline
${L\;''}$ &Linear differential operator.\\
\hline
$F_{s}, C_{s}, B_{s}$ &Fine-structure, Coarse-structure, and Base layers \\
\hline
$w(P, Q)$ & Weighting matrix filter with parameters P and Q.\\
\hline
$\mathcal{V}_{p}$&Visual saliency map.\\
\hline
$H$, $h_n$& Concatenated vector, D dimensional features vector.\\
\hline
$W$, $G_{l+r}$ & Learnable weighted parameter, Graph residual split attention network layer.\\
\hline

\end{tabular}
\end{table}


\section{Related Work}\label{Related work}

The literature encompasses several UAV detection methodologies based on RGB or IR images, leveraging Convolutional Neural Networks (CNN) \cite{10.1145/3582099.3582113, 10.1145/3427796.3428480,10.1145/3583074,10.1145/3579999,10.1145/3376067.3376098,8078541,8204318, 10.1145/3290420.3290426}. In particular, Tian et al. introduced a YOLOv5-based detection paradigm tailored for small UAVs in \cite{10.1145/3582099.3582113}. This model refines detection by designing anchor box sizes, incorporating a convolutional block attention module (CBAM), and revising the loss function. Its prowess is accentuated by its capability to detect UAVs in complex and challenging environments. Similarly, Alsoliman et al. put forth a UAV detection approach that leverages random forest classification, as articulated in \cite{10.1145/3427796.3428480}. This technique discerns patterns in video data to curtail the influx of packets emanating from UAVs. Furthermore, a distinctive method is delineated in \cite{10.1145/3579999}, introducing the notion of a pivot fingerprint model designed for pinpointing anchor packets within video streams for UAV detection. The framework of this model uses a two-tiered feature selection process, with the first phase being model-independent and the second phase being model-dependent.

Lui et al. studied an HR-YOLACT algorithm, which is an amalgamation of HRNet and YOLACT techniques. This model is architected to feature a lightweight prediction head, facilitating the detection of UAVs and extracting their features through instance-based semantic segmentation \cite{10.1145/3376067.3376098}.
Muhammad et al., on the other hand, delved into a transfer learning technique for the identification of UAVs, using both VGG16 and Faster-RCNN \cite{8078541}. Meanwhile, Jihun et al. brought forward an approach in \cite{8204318}, leveraging a Pan-Tilt-Zoom camera system for UAV detection using the Faster R-CNN Inception Resnet algorithm.

Furthermore, Wei et al. tailored the YOLOv3 model combined with transfer learning to detect UAVs using an RGB camera. This integration has further potential for real-time surveillance applications, especially on platforms like the NVIDIA Jetson TX2 \cite{9358449}. In the same way, Reddy et al. showed how YOLOv3 could be used to find UAVs, highlighting how useful it is for effective monitoring in a variety of daytime situations \cite{10149935}. Lee et al. ventured into a machine learning-centric approach, focusing on the identification of UAVs from RGB images. Their system keeps a vigilant eye on surveillance zones, pinpointing and cataloging UAVs using an RGB camera and subsequently determining their geographic position and manufacturer model \cite{8539442}.

In another innovative approach, Wang et al. launched a semi-supervised object detection technique termed Decoupled Teacher. Built on Faster-RCNN's foundation, this method employs unlabeled data with the aim of counteracting the imbalance between foreground and background observed in RGB camera feeds \cite{10065897}. In \cite{9493767}, Basak et al. describe a YOLO-based UAV detection strategy that uses spectrogram images to classify and group spectral instances.

Byunggil and Daegun, in \cite{8627760}, showcased various micro-Doppler signatures of UAVs. These were discerned using the short-time Fourier transform along with the Wigner-Ville distribution, both of which serve as tools to aid in the identification of UAVs. Multiple trainable object detection models were put to the test, comparing them with UAV identification tasks. Complementarily, Suh et al. embarked on research tailored for UAV detection on platforms that are constrained in terms of resources, particularly focusing on video hardware. Their study in \cite{10.1145/3583074} meticulously melded algorithmic optimizations with FPGA hardware, aiming to adeptly scrutinize the intricacies of video streaming. 

Qi et al. put forth a technique tailored to the recognition of consumer-grade UAVs \cite{10.1145/3290420.3290426}, employing static infrared imagery. Their model uses an approach based on importance and integrates basic convolution, adaptive thresholding, linked domain filtering, and SVM-based discrimination. Sun et al., in \cite{9141228}, introduced TIB-Net, a specialized model designed for UAV detection through an RGB camera. This model, uniquely structured with a cyclic pathway, is particularly adept at detecting small-sized UAVs. Augmenting its efficacy, a spatial attention module is integrated, working to pare down data redundancy and extraneous noise.
Further research, as outlined in \cite{9904469,9607468}, showcases efforts wherein various CNN-based strategies are deployed to identify UAVs. These detection tasks are undertaken under a myriad of lighting scenarios, harnessing the power of RGB video feeds. However, despite the extensive study into CNN algorithms tailored for both RGB and IR-based UAV detection, their potency tends to be encumbered. This restriction results primarily from the difficult challenge of backgrounds with a high degree of similarity to the UAVs.

The prevailing models as depicted in the literature predominantly operate on single-stream data, either RGB or IR, as opposed to processing fused optical flow images, where the visual presentation of UAVs and other entities can exhibit substantial variation between day and night vision scenarios. Additionally, the feature quality of the resultant frames is found to be considerably influenced by varying lighting conditions, which, in turn, often exacerbates the challenges due to the heightened similarities between the UAVs and their background environments.  

\section{The Proposed OF-GPRN System}\label{Methodology}
In this section, we study the proposed OF-GPRN system that learns the feature layers while improving the detection accuracy, which is illustrated in Fig. \ref{ModelDesign}. The OF-GPRN is developed to retain and recognize the original structure of the complex day and night vision data. 
\begin{figure*}[htbp!]
    \begin{center}
    \hspace*{-6em}
        \includegraphics[scale=0.18]{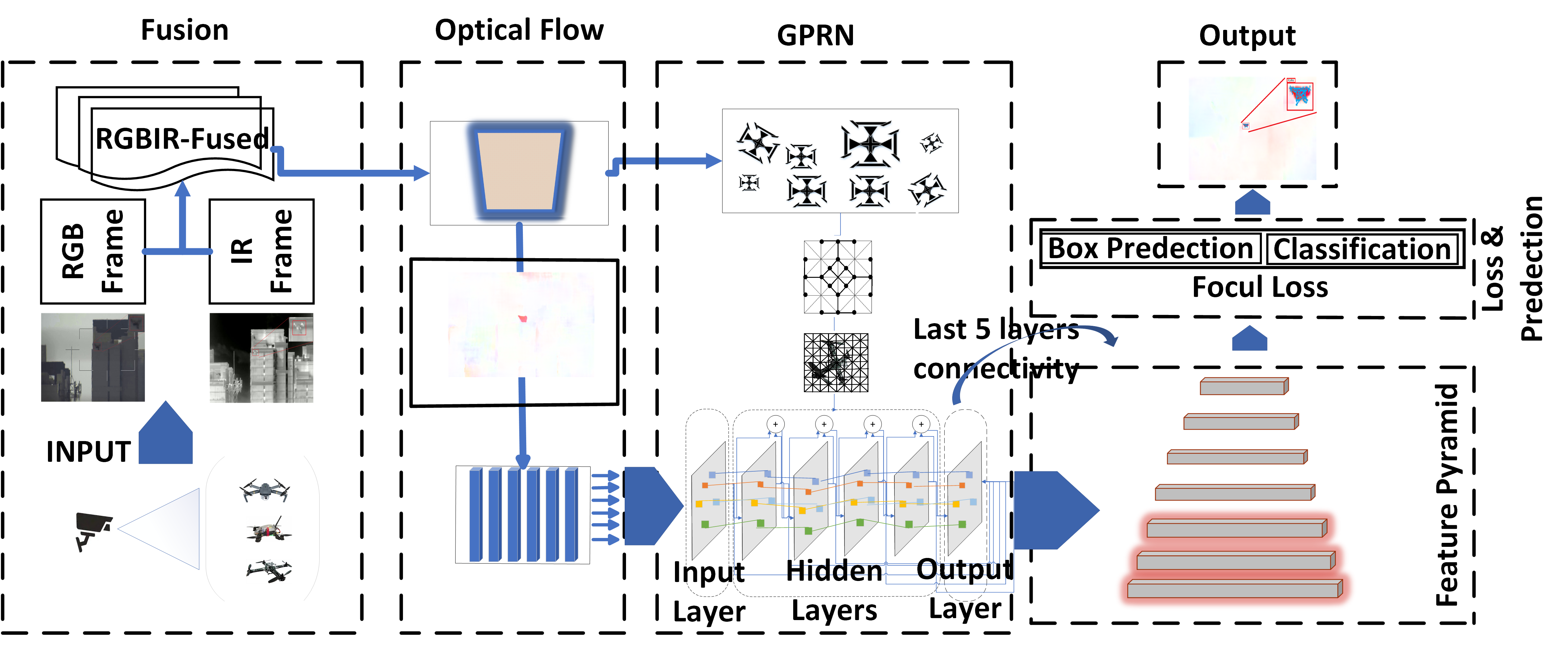}
        \caption{The proposed OF-GPRN system architecture for UAV detection and tracking in day and night vision.} 
        \label{ModelDesign}
    \end{center}
\end{figure*}
\subsection{Proposed System Model}
\subsubsection{RGB-IR Fusion} 
A fusion module is developed in the proposed OF-GPRN to merge two input frames from RGB and IR sources while enhancing the quality of the output frames by reducing noise and adjusting illumination and color. First, multi-level edge preservation filtering is used to separate the input frames into base layers ($B_s$), fine-structure ($F_s$), and coarse-structure ($C_s$). This enables the extraction of fine texture and large texture features at $F_s$ and $C_s$ layers respectively, while contrast and edge details are preserved in the $B_s$ layer. To retain edge information, weighted mean curvature ($W_f$) is applied to each input layer, and a Gaussian filter ($G_f$) is utilized to remove Gaussian noise. The weighting matrix filter $w(P, Q)$ assigns higher weights for center pixels within the square-shaped window, with parameters $P$ and $Q$. A modified Laplacian operator $\mathcal{L}_{(x,y)}$ is defined to obtain high-quality output frames that capture detailed features, such as edges and contours, from both RGB and IR sources \cite{YIN20142274, 9689018}. Thus $\mathcal{L}_{(x,y)}$ is given as 

 ´

\begin{equation}
\label{L_{nsm}}
\mathcal{L}_{(x,y)} = \sum\limits_{P = - p}^p {\sum\limits_{Q = - q}^q w } (P,Q){\left[ {{{L\;}_{(x + P,y + Q)}}} \right]^2}
\end{equation}
Furthermore, the modified Laplacian can be given by ${L\;}_{(x,y)} = |2{L\;''}_{(x,y)}-{L\;''}_{(x-1,y)}-{L\;''}_{(x+1,y)}|+|2{L\;''}_{(x,y)}-{L\;''}_{(x,y-1)}-{L\;''}_{(x,y+1)}|$, where ${L\;''}$ is a linear differential operator that approximates the second derivative at the $F_s$ layer, i.e., ${{L\;''}_{{F_{s}}_{(x,y)}}}$= \{$\frac{\delta^2 F_{s}}{\delta x^2}$ + $\frac{\delta^2 F_{s}}{\delta y^2}$\}. Likewise, $L''_{C_s(x,y)}$ at the $C_s$ layer can also be obtained. In the proposed OF-GPRN, a pulse-coupled neural network with parameter adaptation is used to fuse the $F_s$ and $C_s$ layers while determining the optimal number of features in each layer \cite{8385209}. Specifically, the network evaluates the edge features of the respective layers, with priority given to those extracted from $\mathcal{L}_{{F_{s}}_{(x,y)}}$ if they are more prominent. If the edge features of $\mathcal{L}_{{C_{s}}_{(x,y)}}$ are more prominent, the pixel features extracted from this layer are added to the fused frame.

The fusion of the $B_s$ layers in the RGB and IR frames combines the contrast information from an IR frame with the texture information from an RGB frame. 
In particular, fusing the $B_s$ layers in RGB and IR frames suffers from each pixel contrast and texture information composition \cite{Yan2021}. Therefore, a visual saliency map , denoted by $\mathcal{V}_{p}$, is constructed based on calculating the intensity value of the pixels, which examines the difference between each pixel and all of its neighbors to generate a saliency value \eqref{SaliencySum}. As a result, $\mathcal{V}_{p}$ preserves the contrast and texture features of the RGB or IR frame, improving the quality of the fusion. In \eqref{SaliencySum}, $n$ represents a specific pixel intensity value within the range of 0 to 255 in the frames $B_{s}^{(RGB,IR)}$ (denoted as $N$), and $\mathcal {I}_{n}$ is the number of pixels with similar intensity to $n$ $\forall$ $I=\{i|(x_{n_i},y_{n_i})=n,1\leq i\leq N\}$ in which $i$ is the individual pixel index. The proposed system uses a feature scaling normalization function $S(.)$ to make sure that the frame features fall in the same range. This is due to the fact that different scales in \eqref{SaliencySum} affect many quantitative pixel features.

\begin{equation}
\label{SaliencySum}
\mathcal{V}_{p}= \sum_{p=x,y}^{x_n,y_n} \mathcal {I}_{n}{\left |{ {S\left({{B_{s_{p}}^{(RGB,IR)}}}\right)} - {S\left({B_{s_{x_n,y_n}}^{(RGB,IR)}}\right)} }\right |},
\end{equation}
where $x$ and $y$ are spatial pixel coordinates, and $x_n$ and $y_n$ represent specific pixel coordinates.

For the fusion of the $B_s$ layer, we formulate \eqref{BSFusion} to merge the saliency maps in the RGB and IR frames, which are represented by $\mathcal{V}_{p}^{RGB}$ and $\mathcal{V}_{p}^{IR}$, respectively.

\begin{equation}
\label{BSFusion}
BASE_{fusion} =\frac{{\alpha + \beta}}{2},\end{equation}
where $\alpha = (\mathcal{V}_{p}^{IR} B_{s}^{{\rm{IR}}} + (1 - \mathcal{V}_{p}^{IR} )B_{s}^{{\rm{RGB}}} )$ and $\beta = (\mathcal{V}_{p}^{RGB} B_{s}^{{\rm{RGB}}} + (1 - \mathcal{V}_{p}^{RGB} )B_{s}^{{\rm{IR}}} )$.
Based on the output of the parameters-adaptive pulse coupled neural network and $BASE_{fusion}$, the fusion of the RGB and IR frame can be obtained by applying the inverse multi-level edge preservation filtering.

To monitor the UAV's movement, we used optical flow BRAFT \cite{9495134} to process the frames, which relies on merged frames. an effective approximation of the actual physical motion being projected onto the fused frame, which provides a concise representation of the parts of the frame that are in motion. Integrating spatio-temporal information helps with background elimination. Once the videos are processed, the best next move for the mobile UAV is determined, and routine execution initiates each new iteration of the model.

\subsubsection{Graph Residual Split Attention Network (GRSaN)}
The construction of graph nodes and edges for input images in the proposed OF-GPRN model depends on using region adjacency graphs. Superpixel segmentation approaches, such as SLIC, Quickshift, and Felzenszwalb, are used to precisely split the frames into regions, which then function as nodes in a system. Subsequently, each of these regions connects together on the basis of their adjacency, which leads to the formation of the edges of the graph. The decision of the superpixel segmentation technique plays a role in determining the precision of a generated graph. These nodes and edges are used as inputs for the GCNs model. GCNs are graph-based architectures based on graph nodes and edges $G=(N, E)$. Instead of using conventional convolutional filters, GCNs use graph convolutional filters in each layer of unordered nodes $N$ with edges $E$, and aside from that, GCNs are just like CNN's. Stacks of pointwise nonlinearities in GCNs serve as the building blocks of filters, while stability and permutation equivariance of GCN architectures with good performance are attributed to the graph characteristics \cite{9356126, 9408381}. UAV pixels in the frame are represented as nodes $N=[n_1,....,n_k]\in \mathbb{R}$, with edges $E=[e_1,...,e_k] \in \mathbb{R}$ defining the relationship between the i-th and j-th UAV pixels in the order pair $k=(i, j)$. The vector $H=[h_{n_1}, h_{n_2},...,h_{n_k}]^T \in \mathbb{R}$ concatenates the feature vector $h_n\in \mathbb{R}$ with D-dimensional features of $n$ nodes. Here's how the information from the input layer $(G_{l})$ which is $ConvOper(G_l, W_l)$ and added to and changed in the output layers.
\begin{equation}
\begin{split}
    G_{l+out}= ConvOper(G_{l+out}, W_{l+out})\\
    +\tau (G_{l}+G_{l+1}+...+G_{l+out-1})
\end{split}    
\end{equation}

$W=[W_1,W_2,...,W_{out}]$ is the learnable weighted parameter of the n layers for node aggregation and updating the graph function to compile neighborhood pixel information \cite{9408381}. 

Each $G_{l+r}$ represents the graph residual split attention network layer of the graph residual network \cite{zhang2020resnest} as shown in Figure \ref{Splitattention}. The output of $G_{l+r}$ is given in equation \ref{GRSA}.   
\begin{equation}
    \label{GRSA}
    G_{out}=\sum_{r=1}^R {(G^{Conc}_{l+r})+\tau (G_{l}), r=1,2,3,4, \ldots R\}}
\end{equation}
$G_{l}$ is the input of the graph residual split attention network layer, $G_{out}$ is the output, and $\tau$ is the strided graph convolution or combined graph convolution with max pooling.
If the dimensions of $G_{out}$ and $G_{l}$ equal, then $\tau$ replaced by the identity matrix $({{{\mathbf{I}}}})$ \cite{KimEL17}. 
The GCN main route output is scaled using $\tau$ as a linear projection with the previous input. $\tau$ is scaled with input using a strided graph convolution or a combined graph convolution linear filter with max pooling. As a result, the number of parameters for ResGCN remains constant rather than increasing, as it does for plain GCN or ResGCN without $\tau$. A simple ResGCN without a projection matrix block can add an input channel to an GCN output; however, as the number of layers increased, performance decreased significantly due to shortcut path accumulation. Furthermore, if the input and output have the same dimension, then ${{{\mathbf{I}}}}$ can reduce computational complexity and have the same effect. 

Where $G^{Conc}_{l+r}$ represents the concatenation of the cardinality groups denoted by $(k)$ for each set of hyperparameters $(R)$.  To have a better understanding of each expression, we listed a more in-depth comprehension of each term:
\begin{itemize}
    \item $G^{Conc}_{l+r}$: This notation refers to a specific data structure that results from concatenating multiple groups, where each group corresponds to a different choice of hyperparameters $(R)$.
    \item $\sum_{r=1}^R {(G^{Conc}_{l+r})}$: This expression represents the sum of these concatenated groups. Moreover, it is adding together the information contained in all the different groups. The result is a comprehensive dataset represented as $G_1 + G_2 + \ldots + G_k$, where each $G^{k}\in {R^{N\times D}}$ corresponds to a specific combination of nodes and the cardinality of the set $k$.
    \item $G^{Conc} = H_{G}(G^1, G^2, \ldots, G^k)$: We define a function $H_{G}$ that takes individual groups $G^1, G^2, \ldots, G^k$ as inputs and concatenates their gradients (features). This step ensures that we capture information from all the vertices in each group.
\end{itemize}

\begin{equation}
\begin{split}
\label{GK}
    G^{k}=\sum_{j=1}^k {F((G_{j})+(w^{agg}_{j}))}+w^{update}_{j}\\
\end{split}    
\end{equation}
where, the updated and aggregated learnable parameters are $w^{agg}_{j}$ and $w^{update}_{j}$. Where $F(.)$ is the aggregation function. $w^{agg}_{j}$ compiles information from vertices in the same cardinal $k$'s neighborhood, whereas $w^{update}_{j}$ applies a non-linear function to the aggregated information to compute new vertex representations in cardinal $k$. The $G_{out}$ is processed by global max pooling, followed by batch normalization and ReLU to stabilize the input by softmax and 1x1 convolution, and then transfer for the next layer, as shown in Fig. \ref{Splitattention}.

\begin{figure}[htbp!]
    \begin{center}
    \hspace*{0em}
        \includegraphics[scale=0.11]{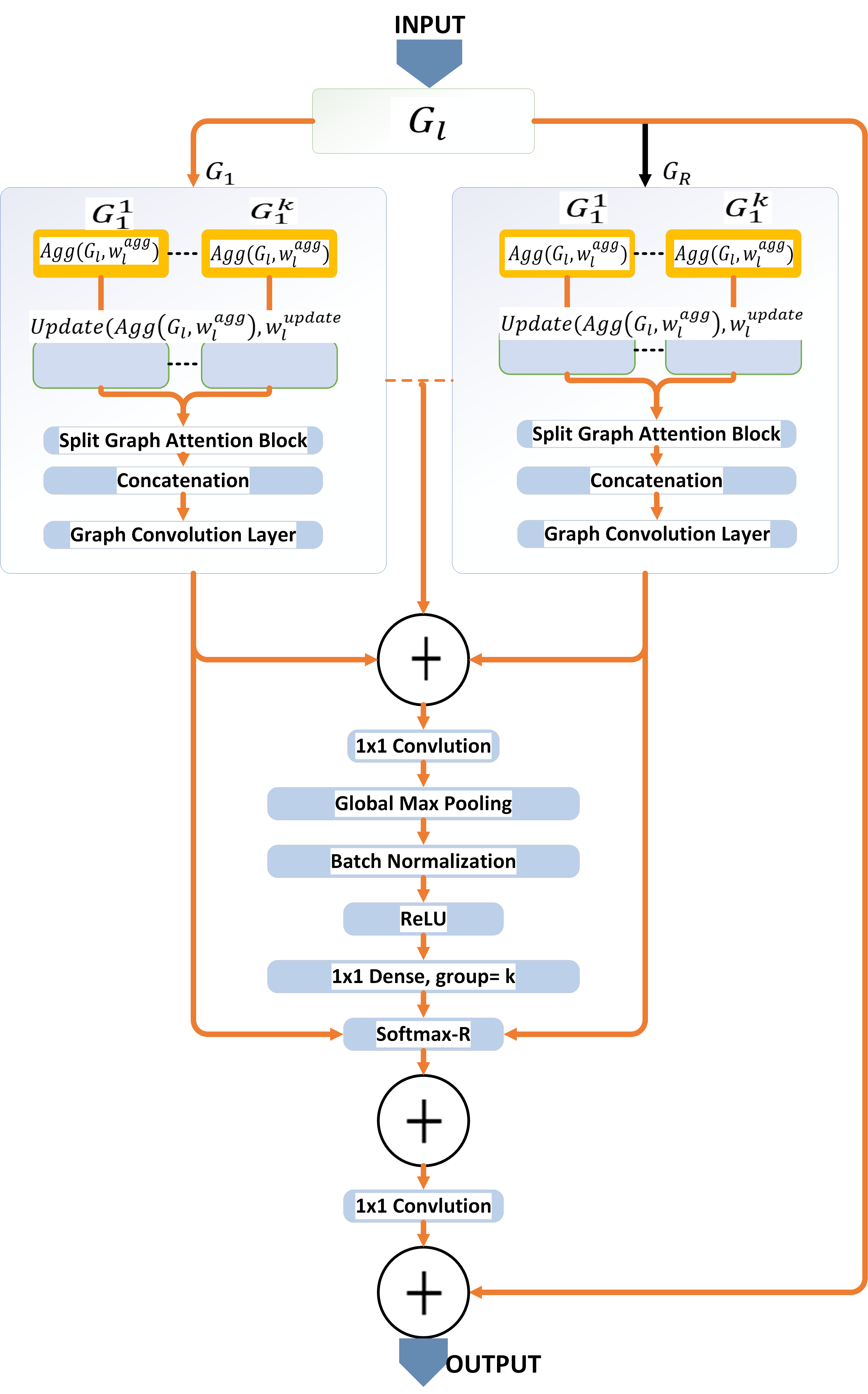}
        \caption{In deep overview of the split graph attention is shown here which represents each layers of the graph convolutional network. }
        \label{Splitattention}
    \end{center}
\end{figure}
\subsubsection{Graph-Pooling Feature Pyramid Network Mapping}
The last layers of the GCN extract high-level features of the input. We selected the last 5 layers for the graph feature pyramid network to map features between GCN and the pooling feature pyramid (PFP) \cite{7005506, 8099589}. We update the last 5 layers of the GCN with a feature pyramid network \cite{GraphFPN}, which has the ability of the superpixel hierarchy to make recursively larger groups of pixels from the smaller features of the last layers (high-level features) and a similarity measure \cite{7423791}. The superpixel hierarchy matches the graph layers, and when moving from one layer of the residual attention GCN last layer to the next, the number of nodes decreases by a factor of 4. Contextual and hierarchical edges are used in different ways in the layer that connects the ancestor and descendant layers, which are called superpixels. Hierarchical edges connect semantic gaps, and contextual edges spread information about the context of the different levels in each layer. The node features used are the same for both hierarchical and contextual layers, but the edges are different for both. Of the last 5 layers, the first and last two are contextual, while the middle layers are hierarchical. Moreover, the learning parameters are different for both and are not shareable. The mapping from GCN to PFP is necessary to transfer the features at multiple scales and make it possible to be in line with the PFP input. Every input from the GCN layer is stride 2, which keeps the input feature from vanishing. An upsampling factor of 2 is applied to features with a higher resolution. Each lateral link combines feature maps of the top-down pathways that are the same size in space. Convolutions are performed on the top-down route feature maps to lower the channel dimensions, and the feature maps from both pathways (input: N/4, N/16, N/64, and N/256; + output: N, N/4, N/16, N/64, and N/256) are combined using element-wise addition. Each combined map is given a 3 x 3 convolution with a factor of 2 to get the final forecast for each layer. For the UAV's position, generative localization of bounding boxes makes a single box with the highest score.

\subsubsection{Loss Function}
In the case of one-stage detection, focal loss is specifically tailored to meet the needs of the user. In the suggested training model, an imbalance between UAVs in the foreground and those in the background could be fixed to put less weight on making accurate predictions. It is common practice to use cross-entropy as a loss function because of its high level of accuracy in comparing the approximation models.
\begin{equation}
\label{crossentropy}
    Cross\ Entropy(_p,_t)=
\begin{cases}    
    -\lambda1_t \ log(p),\ \text{if} \ t=1 \\  
    -\lambda1_t \ log(1-p),\ \text{otherwise} \\
    \text{When},\ t\in\{\pm 1\}, p\in [0,1]
\end{cases}    
\end{equation}

In (\ref{crossentropy}), $t$ is the value of the UAV detection target, and $p$ is a probabilistic estimate of that target value based on the probability distribution. Where $ \lambda1_t $ are the balanced parameters for positive and negative examples; however, it cannot discriminate between simple and challenging cases. The down weight approach requires the modulated focal loss factor $(1-p_t)^\gamma $ for numerical stability.

However, when training naively with (\ref{crossentropy}), the classifier is unable to discriminate between the more accurate candidate and the loose counterpart, resulting in an unanticipated learning scenario as shown in Fig. \ref{Loss_difference}. Because the candidate boxes with more precise locations are suppressed with non-maximum-suppression procedures, this may have a negative impact on performance.
\begin{figure*}[htbp!]
    \begin{center}
    \hspace*{0em}
        \includegraphics[scale=0.8]{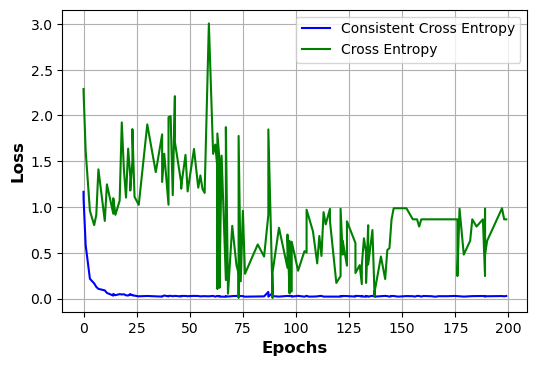}
        \caption{The scaling factor's impact on cross-entropy to automatically downweight loose samples during training.}
        \label{Loss_difference}
    \end{center}
\end{figure*}
As shown in Fig. \ref{Loss_difference}, the consistent cross-entropy loss function is a dynamically scaled cross-entropy loss with the scaling factor determined by the overlap between the current bounding box and the target ground-truth item \cite{8721661}. This scaling factor, intuitively, automatically downweights the contribution of loose samples during training, allowing the model to concentrate on more accurate predictions. The consistent cross-entropy loss function may help train our model to better identify which prediction is the best among numerous clustered choices. So, modulating factors are added to (\ref{crossentropy}) using a consistent cross-entropy loss function \cite{8721661} to accommodate localization quality, and the more precise targets are augmented to reflect it. The (\ref{crossentropy}) updated form is shown in (\ref{crossentropyUpdated}).

\begin{equation}
\begin{aligned}
\label{crossentropyUpdated}
Cross\ Entropy_{(p,t)}= \\
\begin{cases}    
     [-\lambda1_t + \lambda2 (o_{k}-\lambda1)z_{k}]\ log(p),\ \text{if} \ t=1 \\  
    [-\lambda1_t + \lambda2 (o_{k}-\lambda1)z_{k}] \ log(1-p),\ \text{otherwise} \\
    \text{When},\ t\in\{\pm 1\}, p\in [0,1]
\end{cases}  
\end{aligned}
\end{equation}

The $z_k:= 1 (if o_k > \alpha, k=1,...,L)$ represents the candidate box of the targets using IoU overlap for the predicted bounding box, and $o_k$ shows IoU overlapping of the predicted and ground truth bounding boxes. Here $k$ is the location of the UAV in the frames.      
Frames with IoU overlap more than may use the modifying factor in (\ref{crossentropyUpdated}) for more favorable examples, which increases the modifying factor and the loss in proportion to the overlap with ground truth targets. As a result of using cross-entropy, the consistent cross-entropy loss function prioritizes cases with bigger IoU overlaps.

The (\ref{crossentropyUpdated}) is updated in the focal loss as:
\begin{equation}
\label{Focal Loss}
    Focal \ Loss(_p,_t)=-{\alpha_t}\ (1-p_t)^\gamma \ log(p_t)
\end{equation}
The focal loss down-weighted tuning process is dependent on the $\gamma$, and it varies from 0 to 2 to adjust the rate of easy examples. If $\gamma$ is 0, then the focal loss is equal to the cross-entropy. The UAV detection scenarios increase the $\gamma$ up to 2 to obtain the best training result. Moreover, during experimentation, we systematically varied $\gamma$ and observed its impact on the performance of the model. Higher $\gamma$ led to better detection rates for challenging scenarios, such as low contrast or occlusion cases. However, excessively high values might lead to overemphasis on hard examples, potentially causing instability.

\section{Experiments and Performance Analysis}\label{Experiments}
\subsection{Experimental Training}
Extensive experiments on the benchmark UAV catch detection dataset were conducted to evaluate the efficacy of OF-GPRN in enhancing the learning performance for UAV detection with day and night vision cameras. The OF-GPRN model is developed in Tensorflow and trained on a workstation with a GeForce RT 3060. The training procedure used 4 batch sizes. Compared to other optimizers, the Adam optimizer is preferred because of its rapid and intuitive convergence on the best solution. The learning rate decays to $10^{-4}$ and $10^{-6}$ when the value of $\beta1$ is set to 0.9 and when the value of $\beta2$ is set to 0.999. 

In addition, the model is trained for 45 hours and 200 epochs. To build the region adjacency graph and decrease the input size, the frames are converted to super pixels using the algorithms SLIC \cite{6205760}, Quickshift \cite{10.1007/978-3-540-88693-8_52} and Felzenszwalb \cite{Felzenszwalb2004} for OF-GPRN training. Progressive focal loss is used to prevent the unanticipated learning scenario and to discriminate between the more accurate candidate and the loose counterpart. textcolor{blue}{ The comparative performance of UAV detection across various models is shown in Table \ref{experimental_results}. The effectiveness of detection demonstrates substantial variation among models, with the proposed model obtaining the highest performance. The assessed approaches consist of RGBIR-ResGCN, Fusion-ResGCN, OF-ResGCN, and the proposed OF-GPRN. The RGBIR-ResGCN model obtained a mAP of 35.1\% but suffered from a comparatively high loss of 4.253, suggesting that its performance is unstable. Fusion-ResGCN demonstrated an enhanced mAP of 55.5\%. However, there is a 2.147 increase in loss value along with this improvement, primarily as a result of difficulties maintaining background features, which reduces stability. The optical flow-based model OF-ResGCN achieved an mAP of 69.9\%. A loss of 2.042, however, indicates that the model's performance is unsatisfactory. In contrast, the proposed OF-GPRN proved significant improvements, with a remarkable mAP of 87.8\% with a small loss of 0.026. This model showed higher efficiency than other models and also maintained consistent accuracy in its predictions. The significant increase in mAP, along with minimal loss, indicates that OF-GPRN has the potential to be a very effective method for the given task of UAV detection. 

\begin{table}[htbp!]

\centering
\caption{Experimental Results with stability and the models Performance}
\label{experimental_results}
\begin{tabular}{|c|c|c|c|}
\hline
\textbf{Method} & \textbf{mAP (\%)} & \textbf{Loss}& \textbf{Comments} \\
\hline
RGBIR-ResGCN & 35.1 & 4.253 &Unstable \\
Fusion-ResGCN & 55.5 & 2.147 &High loss due to background \\
OF-ResGCN & 69.9 & 2.042 &Lower performance \\
\hline
OF-GPRN (Proposed) & 87.8 & 0.026 &Stable with improved mAP \\
\hline
\end{tabular}
\end{table}    
\subsection{Datasets Pre-processing}
In this study, we employ freely accessible anti-UAV capture video datasets \cite{UAVCatch}. There are a total of 320 clips here, 160 of which are HD videos shot in both standard definition (SD) and high definition (HD) (RGB and IR), which have different variations like backgrounds (cloud, building, mountain, and sea), fast movement, out of focus, and size variations from small to large. The UAVs seen in each video come in a range of sizes, from large to small. We selected UAVs with a size range of 300mm to 1200mm to train and validate our proposed model. Moreover, the UAVs cruise at speeds between 50 and 100 miles per hour and occasionally stop altogether. One hundred validation videos from each stream were used for model training. All except 80 of the videos are used to train the model, while the remaining 20 are used for testing and validation. 

There are a wide variety of variables in the background that may be visible in the video clips, including day and night, lighting conditions, cloudy and clear skies, buildings, and varied degrees of occlusion, as shown in Fig. \ref{fusion_results}. These background variations make it difficult for the detection algorithms to identify the movable object due to low contrast, weak edge details, colour similarities, and texture information. The dataset frames are preprocessed through RGB-IR fusion and an optical flow algorithm to make the movable objects visible.  

\subsection{Effect of Optical Fusion}
The conversion of RGB or IR to optical flow is seen in Fig. \ref{fusion_results}. When compared side by side, all of these frames have distinct levels of performance. A closer inspection reveals obvious artifacts, blurriness, and distinctions in the results of the three columns. When compared to fusion-generated frames, we observe that optical flow fails to inject as many of the bright feature characteristics from the RGB (first column) and IR (second column) frames. The fusion-generated RGB and IR (third column) frames support the optical flow to detect the moveable UAV. Without fusion, the results of optical flow show that the moveable UAV is unidentified due to its high background similarities. Therefore, the spatial properties of the source frames are enhanced in the fusion plus optical flow output frames, which are also free of artifacts, clearer, include more structural details, and have a higher overall visual quality.

\begin{figure*}[htbp!]
    \begin{center}
    \hspace*{1em}
        \includegraphics[scale=0.55]{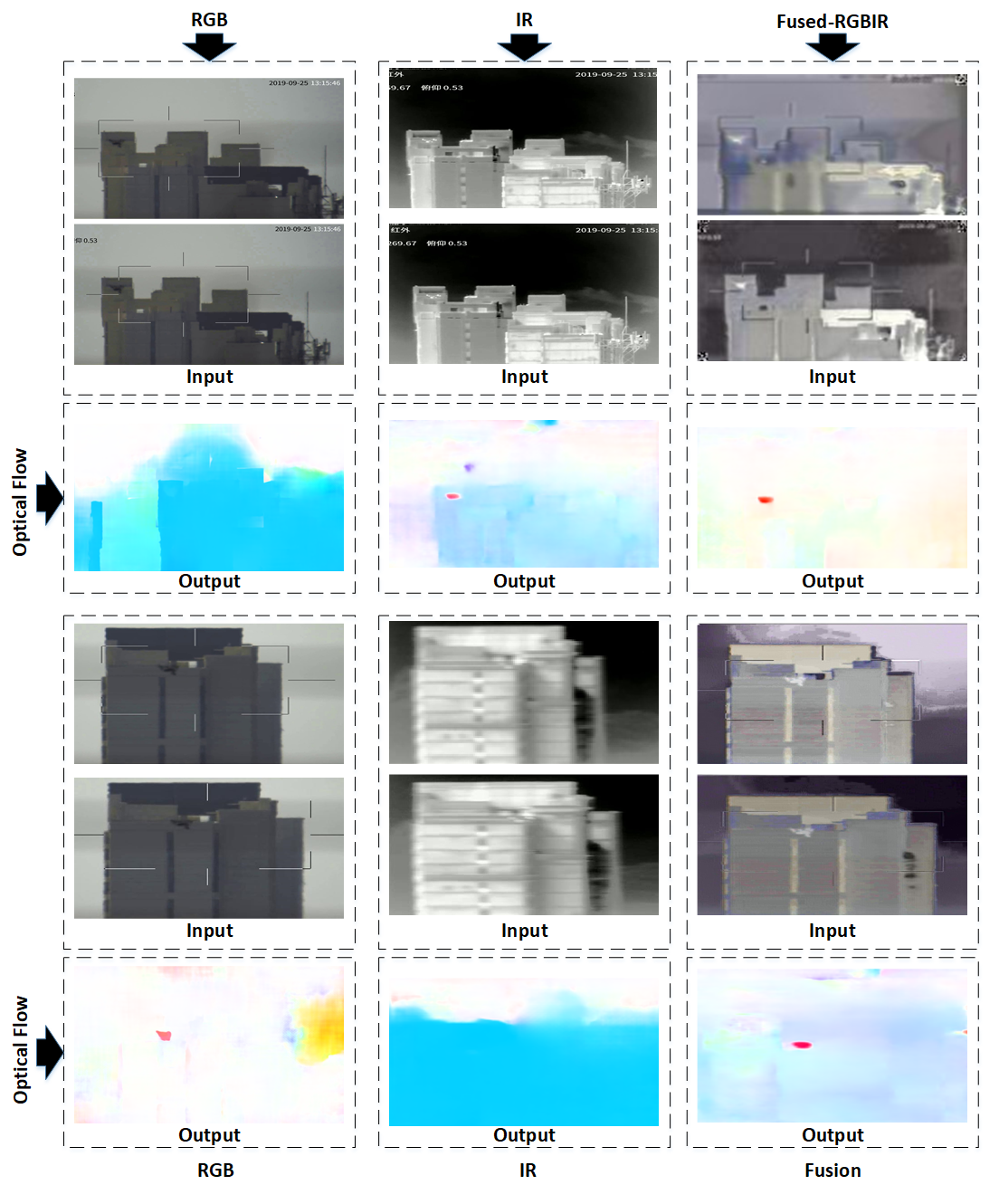}
        \caption{The columns on the far right are RGB frames with object optical flow. The middle column shows the IR frames and optical flow. In contrast, the most right-hand column combines RGB-IR and then displays the effect of UAV optical flow.}
        \label{fusion_results}
    \end{center}
\end{figure*}
\subsection{Effect of Region Adjacency Graphs}
Training the GCN model relies on accurate region adjacency graphs. We can see that the pixel segmentation before the optical, as shown in Fig. \ref{effect_OP_SAG_results} is very complicated compared to the after optical flow. The optical flow for background removal significantly reduces the segmentation of pixels for the region adjacency matrix up to 30\%. Moreover, if the appropriate pixel segmentation procedure is used, the resulting region adjacency graph would be accurate. We used three different superpixel segmentation techniques (SLIC, Felzenszwalb, and Quickshift) to significantly minimize the size of the nodes and edges in the matrix to train the OF-GPRN, as shown in Fig. \ref{effect_OP_SAG_results}. Each of the algorithms identifies regions with similar visual properties. Each frame's features and their associated graph are generated by the adjacency matrix, which also provides concise information regarding the frame's pixels. 

From our experience training the OF-GPRN, we can say that the SLIC algorithm is both fast and space-efficient, and it is able to successfully segment in terms of color boundaries without having to remove the background. However, it recorded the pixels in the background, which makes it less accurate. Additionally, OF-GPRN training using the Felzenszwalb algorithm performs less well due to contrast-based training when it comes to loss minimization. While the Quickshift method is used for the adjacency matrix to achieve OF-GPRN-based promising results of loss 0.026, as shown in Fig. \ref{AfterFusionplusOptical}. Quickshift improves in this regard since it uses hierarchical segmentation computation to separate the image into visually distinct parts. The Quickshift algorithm is used for the proposed model due to its high performance. We achieved the optimal loss of all three algorithms with the parameters mentioned in the Table. \ref{segmentation_summary}. 
\begin{figure*}[htbp!]
    \begin{center}
    \hspace*{0em}
        \includegraphics[scale=0.55]{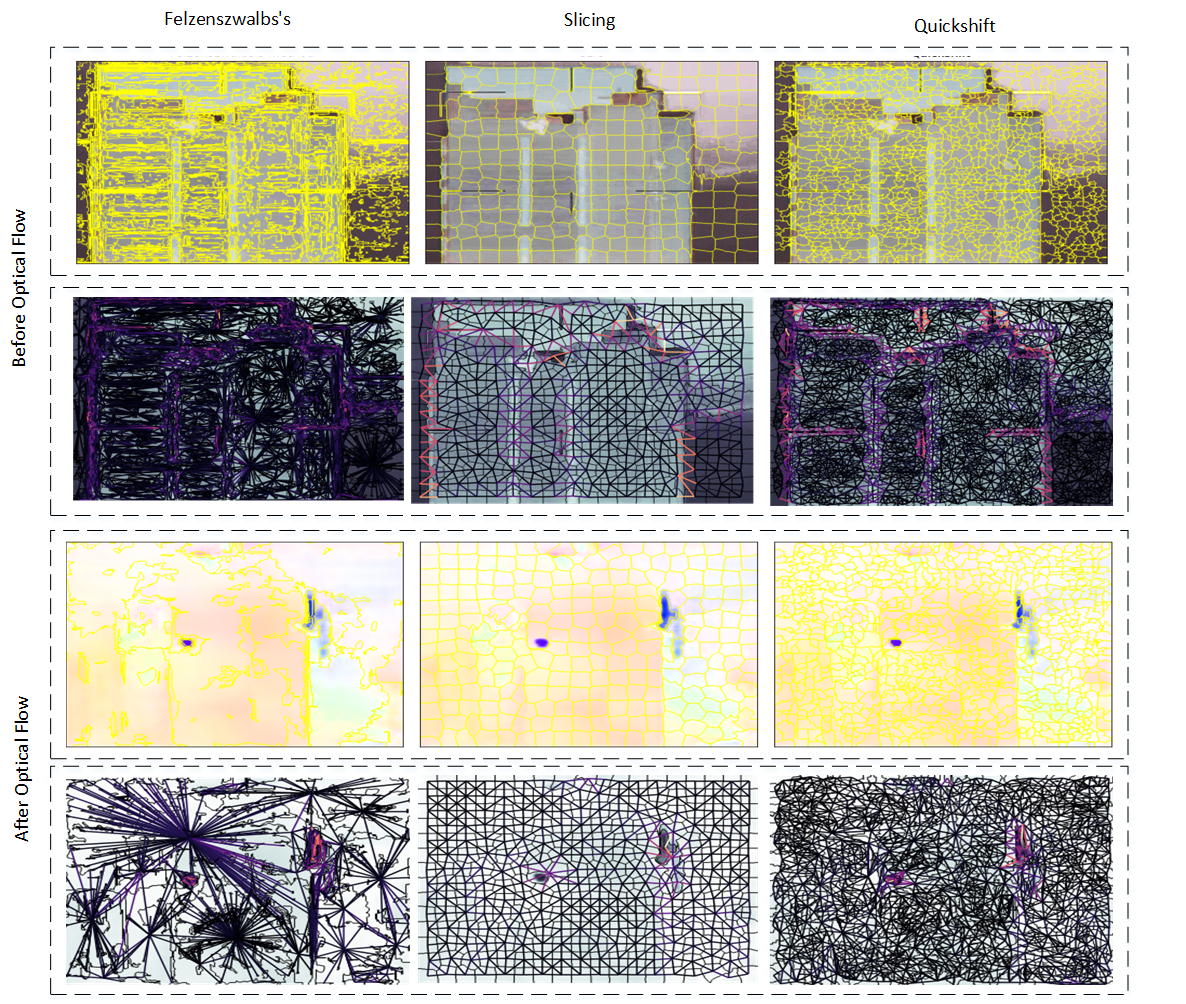}
        \caption{The region adjacency graphs before optical flow and after optical flow have three different segment approaches: The most left column is the Felzenszwalb algorithm performance, the middle column is the SLIC algorithm results, and the most right column is the Quickshift algorithm segmentation for the adjacency matrix.}
        \label{effect_OP_SAG_results}
    \end{center}
\end{figure*}

We use the Quickshift algorithm for exceptional performance after applying optical flow, a method that successfully removes the background. The decision to strategically employ the Quickshift algorithm is based on its notable efficiency in comparison to SLIC and Felzenzswalb. The Quickshift method works better than SLIC and Felzenzswalb in terms of loss efficiency and accurate capture of important features in the image, as shown in Fig. \ref{AfterFusionplusOptical}.
\begin{figure}[htbp!]
    \begin{center}
    \hspace*{0em}
        \includegraphics[scale=1]{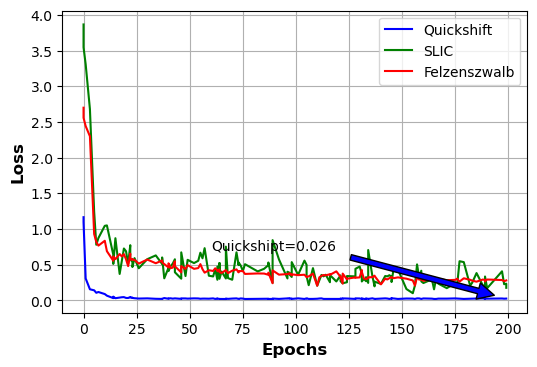}
        \caption{The training losses of the OF-GPRN (based on the Split Attention Network) with three different segment approaches for region adjacency graph algorithms after fusion plus optical flow.}
        \label{AfterFusionplusOptical}
    \end{center}
\end{figure}

\begin{table}[htbp!]
 \caption{Summary of Segmentation Methods}
  \label{segmentation_summary}
  \centering
  \begin{tabular}{|c|c|c|c|c|}
    \hline
    \multirow{2}{*}{\textbf{RAG Methods}} & \multicolumn{4}{c|}{\textbf{Hyperparameters $(P)$ selection and Effect}} \\
    \cline{2-5} & \textbf{$P_1$} & \textbf{$P_2$} &\textbf{$P_3$ } & \textbf{Loss} \\
    \hline
     \multirow{4}{*}{SLIC} & \textbf{Number of Segments} & \textbf{Compactness}& \textbf{Sigma} &\\
    \cline{2-5}
    & 100 & 25 & 0.3 &0.184\\
    \cline{2-5}
    & 1000 &10 & 0.7 &0.235\\
    \cline{2-5}
    & 250 & 20& 0.5&0.176\\
    \cline{1-1}\cline{2-5}
    \multirow{4}{*}{Felzenszwalb} & \textbf{Scale}& \textbf{Sigma}& \textbf{Min size}&  \\
    \cline{2-5}
    &40 & 8& 0.9 & 0.322 \\
    \cline{2-5}
   &100&10 & 0.7 & 0.801 \\
    \cline{2-5}
    &50& 10 & 0.5& 0.281 \\
    \hline
    
    \multirow{4}{*}{Quickshift} & \textbf{Kernel size}& \textbf{Max Dist}& \textbf{Ratio} &  \\
    \cline{2-4}\cline{4-5}
    & 24&8& 0.8 & 0.158 \\
    \cline{2-4}\cline{4-5}
    & 6&4 & 0.2 & 0.059\\
    \cline{2-4}\cline{4-5}
    & 3& 6& 0.5 & \textbf{0.026}\\
     \hline
  \end{tabular}
 
\end{table}

\subsection{Residual Split Attention Network Effect}
Extensive studies, beyond the results of the Quickshift algorithm, are done to illustrate the efficacy of the Split Attention Network in enhancing the learning performance ofGCN with deep architectures.  In Fig. \ref{Different_architectures}, we show how well, without overfitting, the proposed OF-GPRN (based on the Split Attention Network) with a loss of 0.026 with learning rate $10^{-6}$ performs compared to the other ResGCN and observe that the proposed OF-GPRN performs best compared to the RGBIR-ResGCN, Fusion-ResGCN and OF-ResGCN. The RGBIR-ResGCN has very high instability for learning rate $10^{-4}$ and low learning variations during training with learning rate of $10^{-6}$ but has a high loss of 4.253. Even Fusion-ResGCN has a very high loss due to background existence in the frame, which leads to a loss of 2.147. Moreover, OF-ResGCN has a lower loss than RGBIR-ResGCN and Fusion-ResGCN; however, it is still unstable, and the loss ends up at 2.042.

\begin{figure}[htbp!]
\centering
\hspace*{-3.5em}
\subfloat[]{\includegraphics[width=3.1in]{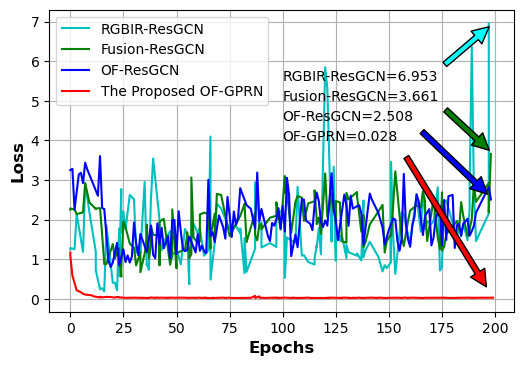}} 
\subfloat[]{\includegraphics[width=3.1in]{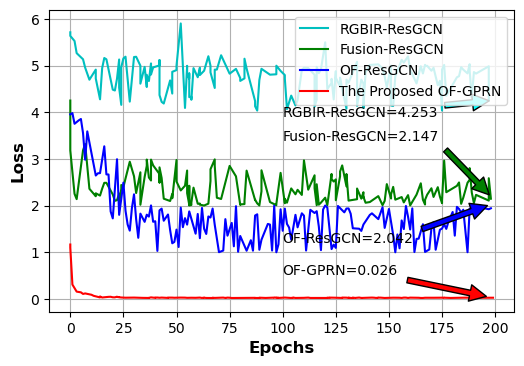}}
\caption{The training loss of the four different graph residual networks is: The cyan line represents the ResGCN with RGB-IR, the green line shows the fusion-based ResGCN, the blue line represents the optical flow fusion ResGCN, and the red line represents the proposed optical flow-assisted graph pooling residual network (OF-GPRN). Moreover, (a) is trained with a learning rate of $10^{-4}$ and (b) is trained with a learning rate of $10^{-6}$.} 
\label{Different_architectures} 
\end{figure}

\begin{figure}[htbp!]
\centering
\hspace*{-3.5em}
\subfloat[]{\includegraphics[width=3.1in]{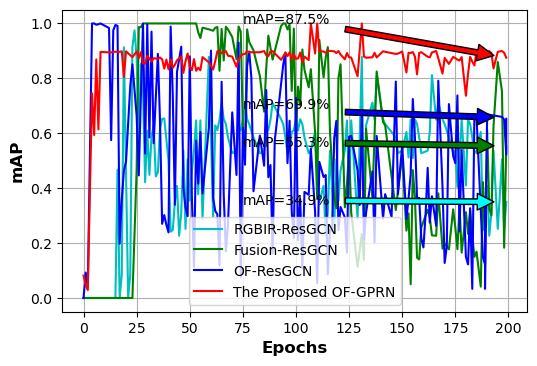}} 
\subfloat[]{\includegraphics[width=3.1in]{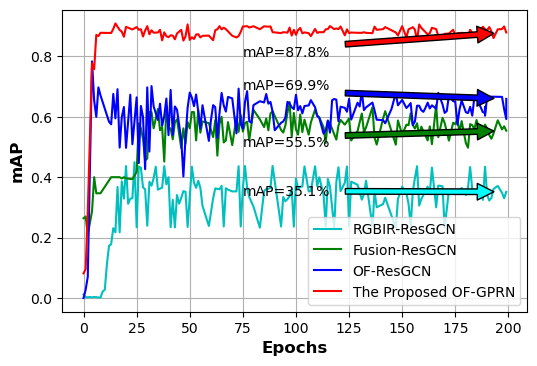}}
\caption{The mean average precision (mAP) performance of the proposed OF-GPRN model vs. the RGBIR ResGCN, Fusion ResGCN, and optical flow ResGCN. Moreover, (a) is trained with a learning rate of $10^{-4}$ and (b) is trained with a learning rate of $10^{-6}$.} 
\label{mAP} 
\end{figure}

\begin{figure}[htbp!]
    \begin{center}
    \hspace*{0em}
        \includegraphics[scale=1]{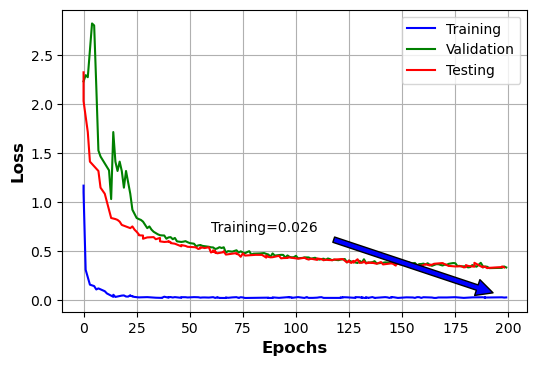}
        \caption{The proposed OF-GPRN model performance in the form of training, validation, and testing losses uses Quickshift for the region adjacency graphs. Moreover, advanced focal loss is used to train the proposed model. }
        \label{Training_Val_Testing}
    \end{center}
\end{figure}
Fig. \ref{Training_Val_Testing} shows both the optimal training loss score of 0.026 and the actual validation loss of 0.331 and the testing loss of 0.35. Moreover, Fig. \ref{mAP} presents the results of the qualitative mAP evaluation with learning rates of $10^{-4}$ and $10^{-6}$ and makes a comparison between the OF-GPRN method and the RGBIR-ResGCN, Fusion-ResGCN, and OF-ResGCN. The OF-GPRN model that has been proposed offers several benefits when it comes to the detection of UAVs. It is clear to observe that the proposed OF-GPRN has a higher mAP with stability during training for different UAV sizes compared to the RGBIR-ResGCN, Fusion-ResGCN, and OF-ResGCN. Moreover, the proposed OF-GPRN achieved 87.8\% more than RGBIR-ResGCN, Fusion-ResGCN, and OF-ResGC, which only reach 35.1\%, 55.5\%, and 69.9\% and are unstable while the model is being trained.

As shown in the results of the experiments in Table \ref{Performance_com} clearly shows that the mAP of the proposed OF-GPRN is much better when it is made up of RGBIR-fused optical flow data. Our proposed OF-GPRN model achieves a highly remarkable mAP of 87.8\%. The OF-GPRN model outperforms previous models, such as the Hybrid-DL, which obtained a mAP of 68.1\%, and the EfficientDet model, which achieved 67.3\% mAP. Importantly, our model shows higher accuracy compared to the DETR model from anti-UAV \cite{10.1007/978-3-030-58452-8-13}, which attained a mAP of 83.2\%. The anti-UAV DETR model, EfficientDet, and Hybrid-DL with the Fusion method all have different performance metrics, as shown in the Table. \ref{Performance_com}. Anti-UAV DETR model has a high precision of 87.22\%, whereas EfficientDet has a similar but slightly lower precision of 71.11\%. Hybrid-DL lies somewhere in the middle, with a precision of 73.55\%. In terms of recall, the anti-UAV DETR model is at 79.61\%, followed by Hybrid-DL at 67.14\% and EfficientDet at 65.33\%. The F1 score shows that EfficientDet scored 68.10\%, Hybrid-DL scored 70.12\%, and anti-UAV DETR model scored 83.24\%.

EfficientDet and Hybrid-DL come in second and third, with the anti-UAV DETR model achieving the highest mAP at 83.2\% and 67.3\%, respectively. In terms of computing efficiency, EfficientDet has the lowest inference time of 102 ms, followed by Hybrid-DL at 107 ms and anti-UAV DETR model at 140 ms. Despite the longer inference time, the proposed model OF-GPRN, outperforms its competitors in precision 91.36\%, recall 89.52\%, and F1 score 90.43\%, indicating a favorable trade-off between performance and computational cost. The proposed OF-GPRN model has a slightly longer inference time of 250 ms but surpasses its competitors in terms of precision, proving its superiority in the application of UAV detection.

As shown in the results of the experiments in Table \ref{Performance_com} clearly shows that the mAP of the proposed OF-GPRN is much better when it is made up of RGBIR-fused optical flow data. Our proposed OF-GPRN model achieves a highly remarkable mAP of 87.8\%. The OF-GPRN model outperforms previous models, such as the Hybrid-DL, which obtained a mAP of 68.1\%, and the EfficientDet model, which achieved 67.3\% mAP. Importantly, our model shows higher accuracy compared to the anti-UAV DETR model, which attained a mAP of 83.2\%.

Upon analyzing the training performance, our proposed approach shows an incredibly small training loss of 0.026. In comparison, the hybrid-DL model has a training loss of 1.13, the EfficientDet model has a loss of 1.20, and the anti-UAV DETR model has a training loss of 0.11. The significant decrease in training loss for our model shows its high efficiency and efficacy in acquiring features from the data. The results highlight the importance of using RGBIR-fused optical flow data to improve the overall performance of the proposed model compared to other methods.
\begin{table}[htbp!]
\centering
\caption{Performance comparison on detection models.}
\label{Performance_com}
\hspace*{-6em}
\begin{tabular}{|c|c|c|c|c|c|c|c|}
\hline
\textbf{Method} &\textbf{Precision } &\textbf{Recall} &F1 Score & \textbf{mAP (\%)} & \textbf{Loss}& \textbf{Inference} &\textbf{Param(M)}\\
\hline
Anti-UAV (DETR) \cite{uavbenchmark}+ Fusion&87.22&79.61&83.24& 83.2 & 0.11 & 140ms&41\\
EfficientDet \cite{9156454}+ Fusion &71.11 &65.33&68.10&67.3 & 1.20 &\textbf{102ms} &\textbf{17}\\
Hybrid-DL \cite{9750351}+ Fusion &73.55&67.14&70.12&68.1 & 1.13 & 107ms&19\\
\hline
OF-GPRN (Proposed) &\textbf{91.36} &\textbf{89.52}&\textbf{90.43}&\textbf{87.5} & \textbf{0.026} & 250ms&31.2\\
\hline
\end{tabular}
\end{table} 

Upon analyzing the training performance, our proposed approach shows an incredibly small training loss of 0.026. In comparison, the hybrid-DL model has a training loss of 1.13, the EfficientDet model has a loss of 1.20, and the anti-UAV DETR model has a training loss of 0.11. The significant decrease in training loss for our model shows its high efficiency and efficacy in acquiring features from the data. The results highlight the importance of using RGBIR-fused optical flow data to improve the overall performance of the proposed model compared to other methods. 

Moreover, the experiment rigorously examined the entire processing time for the fusion method, Quickshift algorithm, and Graph Convolutional Network (GCN). More precisely, the fusion technique required a processing time of 17 milliseconds, the Quickshift algorithm cost 130 milliseconds, and the GCN algorithm utilized roughly 103 milliseconds. The total processing time for all three parts is 250 milliseconds. This detailed timing study provides valuable insights into the computational efficiency of each part, facilitating the evaluation of their respective contributions to the overall time required for processing. 

The proposed OF-GPRN relies on optical flow-assisted graph-pooling and residual networks to enhance UAV detection. One limitation is its sensitivity to changes in environmental conditions, such as abrupt rain, which creates noises in the image backgrounds, and it would be challenging for the optical flow to identify the UAVs. In our future work, we could explore strategies to improve the algorithm's robustness under varying conditions.

Moreover, the effectiveness of the OF-GPRN system is demonstrated using a benchmark UAV catch dataset. However, the algorithm's performance may be influenced by variations in different UAV shapes, tiny sizes, and very fast motion patterns not covered comprehensively in the training data. Discussing strategies for handling diverse UAV scenarios and potential limitations related to data variations would strengthen the paper.
The proposed system is evaluated on a specific dataset, and its generalization to new data environments may be a concern due to the feature variations. The OF-GPRN system model uses a multi-stage approach that includes fusion RGB and IR signals, optical flow processing, and the Graph-Pooling Residual Network. While these stages contribute to improved detection rates using high-computational real-time ground-based static servers, the OF-GPRN system model may face computational complexities, especially in real-time applications with onboard devices. Addressing the trade-off between accuracy and computational efficiency could be an important consideration.

\section{Conclusion}\label{Conclusion}
In this paper, we propose a new OF-GPRN system to enable the precise detection of UAVs in dual day and night visions that suffer from time-varying lighting conditions and high background similarities. The proposed OF-GPRN system incorporates optical fusion techniques and effectively eliminates extraneous backgrounds, resulting in enhanced clarity of RGB/IR imaging. Moreover, the GRSaN is extended in the OF-GPRN system to stabilize the learning capability, improve feature learning during training, and reshape the UAV. In addition, the OF-GPRN system extracts the pixels-to-node representation of the adjacency matrix to achieve an accurate graph with information about the pixels of the conspicuous frame and has the precise learning ability to capture correlations among the pixels in the fused images to identify the suspicious UAV. Experimental results show that our proposed OF-GPRN system achieves an impressively low loss of 0.026. Compared to previous ResGCN object detectors, which recorded an mAP of 69.9\%, the OF-GPRN system delivers superior performance, reaching an mAP of 87.8\% on the demanding RGB-IR-based benchmark UAV catch dataset. Moreover, optimization of the model should be the objective of future studies for onboard real-time system models like UAVs. In addition, optical fusion limits the present study to static cameras; we aim to improve the model for implementation with moving objects, such as autonomous vehicles and UAVs. Research on its possible applications in different environmental circumstances, such as rain and different types of UAVs, is required. Improving accuracy and efficiency requires testing on more and more diverse datasets.


\section*{Acknowledgment}
This work was supported by the CISTER Research Unit (UIDP/UIDB/04234/2020) and project ADANET (PTDC/EEICOM/3362/2021), financed by National Funds through FCT/MCTES (Portuguese Foundation for Science and Technology). Also, this article is a result of the project NORTE-01-0145-FEDER-000062 (RETINA), supported by Norte Portugal Regional Operational Programme (NORTE 2020), under the PORTUGAL 2020 Partnership Agreement, through the European Regional Development Fund (ERDF).



\bibliography{Bibfile}

\end{document}